\documentclass[runningheads]{llncs}
\usepackage[T1]{fontenc}
\usepackage{graphicx}
\usepackage[hyphens]{url} 
\usepackage{authblk}

\begin{document}
\bibliographystyle{splncs04}

\title{Problem Classification for \\ Tailored Helpdesk Auto-Replies}
\author{Reece Nicholls, Ryan Fellows, Steve Battle, and Hisham Ihshaish}
\institute{Computer Science Research Centre (CSRC),\\ University of the West of England (UWE), Bristol, UK}
\maketitle

\begin{abstract}
IT helpdesks are charged with the task of responding quickly to user queries. To give the user confidence that their query matters, the helpdesk will auto-reply to the user with confirmation that their query has been received and logged. This auto-reply may include generic `boiler-plate' text that addresses common problems of the day, with relevant information and links. The approach explored here is to tailor the content of the auto-reply to the user's problem, so as to increase the relevance of the information included. Problem classification is achieved by training a neural network on a suitable corpus of IT helpdesk email data. While this is no substitute for follow-up by helpdesk agents, the aim is that this system will provide a practical stop-gap. 
\end{abstract}
\section{Introduction}
The university IT helpdesk system studied in this research uses a ``call'' ticket system to manage requests and incidents from students and staff. These tickets are usually generated by requests from self-service web services, email, phone, web chats, and walk-in advice points. Currently, a generic acknowledgement email is sent back with the most frequently asked issues (for example, guidance on resetting passwords). Ultimately, these tickets are processed by human agents who will assign the ticket to the relevant agent or team for resolution.

Many of these tickets are requests for information, which may already be available on help pages or FAQ pages. An automated system that can process these emails and automatically reply with the appropriate information may provide useful interim support, and in the best-case scenario may even resolve the query raised by the user.

The dataset used in this paper is a real-world corpus of IT helpdesk interactions between University students \& staff, and helpdesk assistants. This dataset consists of 600 email threads and 5697 follow-up emails, including the direction (incoming or outgoing), subject line, email body and time-stamp. A small selection of emails can be seen in table \ref{tab:emails}. For this study we use only incoming emails, and for simplicity the subject line and email body have been concatenated. All punctuation has been removed and the time-stamp has been discarded. The interactions consist of a wide range of natural language queries that helpdesk staff must try to resolve. The dataset is fully anonymised at source, so that the identity of individuals is not revealed. In addition we attempt to remove any additional contextual information about the locale that could provide clues about their identity.

\begin{table}
\centering
\begin{tabular}{p{11cm}}
``creative cloud Hi I can no longer load any of the adobe products on my PC as It says licence has expired Is this a known issue''\\ \vspace{6pt}
``Solidworks Good Morning I tried installing solidworks off the appsanywhere cloudpaging player but whenever i go to launch it it gets removed straight away and i cant install it Please advise Cheers'' \\ \vspace{6pt}
``Hi I tried to access my blackboard and it kept giving me an error message Hi I tried to access my blackboard and it kept giving me an error message''\\ \vspace{6pt}
``Forgotten password Hi Im having trouble logging in to my student account Ive forgotten my login and password Can you help please Kind regards'' \\ \vspace{6pt}
``Temporary login ID for visitor needs wifi access would like to process a temporary ID for external visitor that needs to connect to wifi Event is tomorrow but shehe may come again after lockdown''\\
\end{tabular}
\vspace*{12pt}
\caption{Examples of uncategorised emails from the dataset. Each one is a concatentation of the subject line and email body, anonymised, and with punctuation removed.}
\label{tab:emails}
\vspace{-24pt}
\end{table}

\section{Natural Language Processing}
Natural Language Processing (NLP) is a field of Computer Science where the goal is to make machines understand our spoken and written languages. NLP must cope with the complexity of grammar, syntax, and vocabulary of languages, as well as the semantics of words used in sentences \cite[p.22]{Ganegedara2018}. Humans are able to read an email and determine which category the email falls into, such as problems with WiFi or passwords etc. Natural Language Processing is used to train a neural network on a corpus of email content, identified as \emph{patterns}. Supervised learning is the task of learning a function that maps an input to an output based on a sample of input-output pairs \cite{Norvig2010}. Every email in the dataset is labelled with the category it belongs to. The training algorithm uses this to learn to classify new incoming queries.

However, the raw data comprises uncategorised emails, so establishing categories for supervised learning is problematic. It's possible to use clustering methods, but without access to a `ground truth', the resulting categories must still be verified by helpdesk agents. Methods like Latent Dirichlet Allocation (LDA) \cite{Blei2003} produce a statistical model that allows sets of observations to be explained by a number of \emph{unobserved} categories, known as a \emph{topic model}. LDA can be used to cluster data, using the presence of words as weighted signifiers of any given category. It enables the top-ranking words for any given category to be listed explicitly, providing the means for users to inspect, verify, and even revise the resulting clusters manually.

\section{Methodology}

 This study side-steps the issues around the initial data categorisation, assuming instead that users are directly able to define categories using a set of keywords, such as may be produced by LDA. Emails in the raw data that match any of these keywords are assigned to that category to produce the training set. The categories and their associated keywords are ordered so that any given email is assigned only to the first matching category, though we strive (manually) to minimise the overlap. This provides a convenient method of categorisation, and because it is user-defined, establishes a ground-truth for supervised learning.

The five categories used in this study, and the keywords associated with them are listed in table \ref{tab:keywords}. Using these keywords to select data and assign a category produces a smaller subset of 265 emails used for training and testing as described in section \ref{training}.

\begin{table}
\centering
\begin{tabular}{rp{9cm}}
  \textbf{Adobe} & \{ `Adobe', `Creative Cloud' \} \\
  \textbf{AppsAnywhere} & \{ `AppsAnywhere', `license', `cloudpaging'\} \\
  \textbf{Blackboard} & \{ `blackboard' \} \\
  \textbf{password} & \{ `password', `mfa', `multifactor', `multi-factor',`multi factor' \} \\
  \textbf{WiFi} & \{ `wifi', `wi-fi', `network', `eduroam', `connection', \newline `internet', `remote access' \} \\
\end{tabular}
\vspace*{12pt}
\caption{Five user-defined categories and their associated keywords}
\label{tab:keywords}
\end{table}

This keyword based approach to defining categories begs the question, ``why not use the keywords directly as a classifier, bypassing the neural network?'' This is a valid approach, so in the evaluation section we compare the neural network solution with a simple classifier based on a \emph{decision tree} -- as a baseline estimator -- trained on a binary vector indicating the presence or otherwise, of each keyword in any given email. This approach echoes the way we search for the samples in the first place, based on keyword matching.

\subsection{Training/test data split}\label{training}
The data is split into training data and testing data, and for these experiments an 80\% training to 20\% test ratio is used throughout. The available data is divided into a training set containing 212 emails, leaving 53 emails for testing. Because we have multi-category data, \emph{stratified sampling} is used to split each category proportionally. This minimises sampling bias in the training set and helps to create a test set which best represents the entire population of data \cite{Menon2020}.

\subsection{Data augmentation}
Following the training/test data split, the training data contains 212 different emails used for training the model. This is a relatively small dataset which is at risk of overfitting by the learning algorithm. To minimise overfitting and increase the accuracy of the model, we use a process called \emph{data augmentation} which Shahul \cite{Shahul2021} describes as a way of generating brand new sentences to train the model. By preventing overfitting, the model is encouraged to generalise \cite{Wang2017}. We use a technique known as \emph{synonym replacement} \cite{Nayak2022} which generates new sentences from existing sentences, effectively re-expressing them.

Not only do we have limited data, but the data is imbalanced in terms of the number of samples in each category. This class imbalance \cite{lopez2021} may create a learning bias towards the categories with more emails. We use a form of \emph{Stratified} Data augmentation that allows us to re-balance the training data so that each category receives equal representation. The email data is augmented so as to present 200 emails per category, which involves augmenting some categories more than others. This produces a training set with 1000 samples, distributed equally over the five categories.

In the evaluation section, we compare classifiers with and without data-augmentation to see if the expected improvement is present.

\subsection{Stemming and stop-word removal}
Before training the model, the data must first be prepared for processing. Word tokenisation chops up a sentence into individual tokens, stripping out punctuation leaving only the individual words or terms of a sentence \cite{Manning2008}.
After tokenisation, we take each word (or token) in the patterns and perform \emph{lemmatisation}, converting it into its equivalent base word \cite{Manning2008}. The process of lemmatisation allows for greater prediction accuracy as it removes all the various tenses and combinations of the same base word. For example, the words `changed', `changes' and `changing' would all be lemmatised to their base word, `change'. This enables the model to generalise, as a model trained on the word `change' would correctly recognise the word `changing' in an email, because they share a common base.

After lemmatisation, the final pre-processing step is to remove stop-words. Stop-words are common words which provide little value to a computer in representing a sentence \cite{Manning2008}. Examples of English stop-words are `and', `are', `from', `he', `is', `the', `it', `the' and `to'. The sentence, ``the internet is not working'' after stop-words are removed is reduced to ``internet not working'', and ``internet not work'' when combined with lemmatisation; both preserving the underlying semantics and both providing a simple sentence representation for supervised learning.

\section{Problem Classification}
Classification is the process of assigning items to a limited set of categories \cite{Alpaydin2014}. Given the text of a user query, we must classify it one of a predefined set of categories, each one of which is associated with a different response that to be included in the helpdesk auto-reply.

The bag-of-words model (BoW) is commonly used for text classification problems, and is the method selected for this research. The frequency of occurrence of each word is represented as a vector and used for training a classifier \cite{McTear2016}. Each email is represented as a bag, or \emph{multiset}, of its words, disregarding grammar and word order, but retaining the number of occurrences. 

This bag of words representation is then used to train the neural network model. We build a sequential model, conventionally known as fully connected feed-forward neural net, comprising a stack of three main layers, where each layer has exactly one input tensor and one output tensor. The input layer is a tensor, a vector representing the bag of words. We allocate 40 units to a single intermediate (hidden) layer. The number of hidden units is determined by starting low, and then incrementing this by stages until no appreciable improvement in accuracy is seen during the learning phase. Finally we have an output layer with one unit for each output category. This \emph{softmax} layer converts the output scores to a normalized probability distribution. Dropout layers, sandwiched between these principal layers, randomly clear their inputs to minimise over-fitting.

The number of epochs – the number of cycles through the training set – is set to 50 epochs, as the improvement in accuracy during the learning phase shows no appreciable improvement beyond this point.

\section{Solution}
Our solution uses the NLTK  library (natural language toolkit), NumPy  and TensorFlow  to perform the machine learning and language processing required to train the model and classify emails. The NLPAug  library is used to perform data augmentation on the training data. For the model evaluation, we use scikit-learn, Pandas and matplotlib  libraries. Following data preparation, the neural network is built using TensorFlow. The result is a trained model used to classify emails into their respective categories.

The classifier is based on the trained model, and uses the NumPy and TensorFlow libraries to predict which category the email belongs to. During evaluation, the classifier tests the trained model against the test data. For the application of this model to incoming emails, currently in prototype, the classification model is applied to individual emails and the response associated with the predicted category is embedded within a generic auto-reply email template. If the prediction error is below a certain threshold (25\%) the system reverts to a standard auto-reply, rather than risk sending out irrelevant information.

\section{Evaluation}

\begin{figure}[t]
\centering
\resizebox{12cm}{!}{\includegraphics{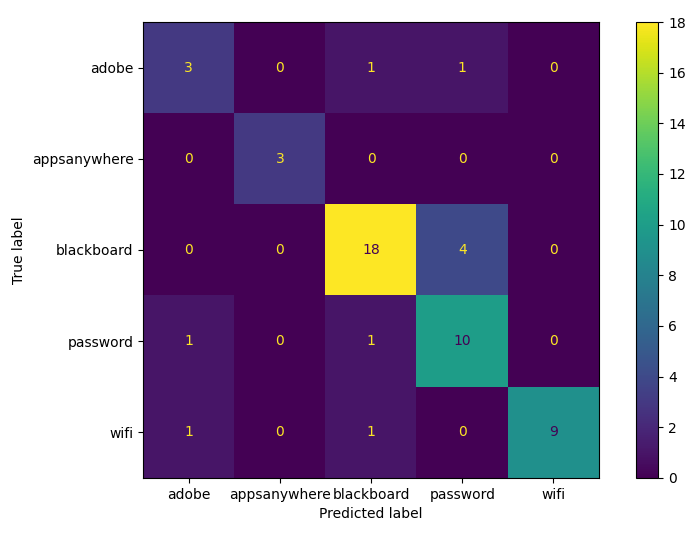}}
\caption{Confusion matrix for neural network without data augmentation.}
\label{fig:confusion_nn_unaugmented}
\end{figure}

\begin{figure}[t]
\centering
\resizebox{12cm}{!}{\includegraphics{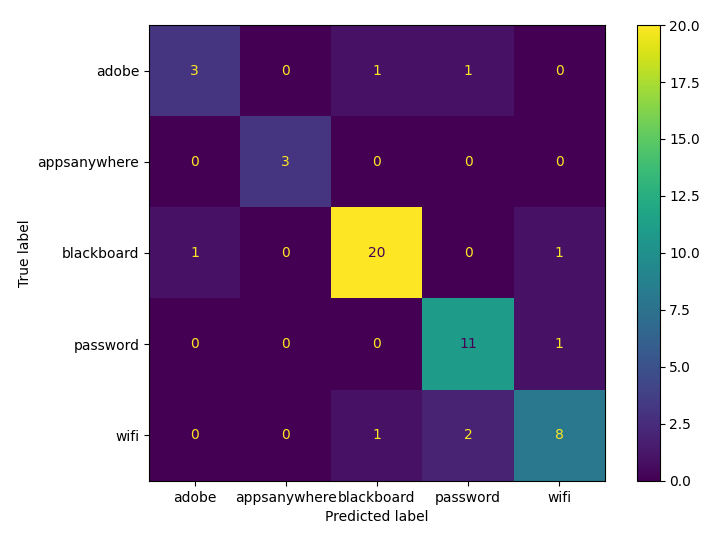}}
\caption{Confusion matrix for neural network with data augmentation.}
\label{fig:confusion_nn_augmented}
\end{figure}

Data augmentation is applied not only to reduce over-fitting, but also to re-balance the training data. We would expect to see improved results. Figure \ref{fig:confusion_nn_unaugmented} shows a \emph{confusion matrix} for the neural network trained on the data without augmentation. A confusion matrix is a table showing the performance and quality of the model, with each row representing the true categories, and the columns representing the predicted categories \cite{Powers2007}. It literally allows you to see if the model is confusing one category with another. A similar confusion matrix for the neural net trained on the augmented data can be seen in figure \ref{fig:confusion_nn_augmented}. The results reported here reflect the median accuracy achieved over a small number of test runs.

%
%

\begin{table}
\centering
\setlength\tabcolsep{4pt}
\begin{tabular}{r|cccc}
& \textbf{precision} & \textbf{recall} & \textbf{f1-score} & \textbf{support}\\
\hline\\
\textbf{adobe} & 0.60 & 0.60 & 0.60 & 5 \\
\textbf{appsanywhere} & 1.00 & 1.00 & 1.00 & 3 \\
\textbf{blackboard} & 0.86 & 0.82 & 0.84  & 22 \\
\textbf{password} & 0.67 & 0.83 & 0.74 & 12\\
\textbf{wifi} & 1.00 & 0.82 & 0.90 & 11\\
\\
\textbf{accuracy} & & & 0.81 & 53\\
\textbf{macro avg} & 0.82 & 0.81 & 0.82 & 53\\
\end{tabular}
\vspace*{12pt}
\caption{Classification report for neural network without data augmentation.}
\label{tab:classification_report_nn_unaugmented}
\end{table}

It's possible to make out greater number of \emph{true positives} in the leading diagonal of the confusion matrix for the model trained on the augmented data. The differences are clearer in the classification reports for each in tables \ref{tab:classification_report_nn_unaugmented} and \ref{tab:classification_report_nn_augmented}. For the model without data augmentation we see lower overall accuracy and lower average precision.

The accuracy is the number of correct predictions divided by the number of samples, expressed as a percentage. The neural net trained on the augmented data set has an overall accuracy of 85\%, compared to 81\% accuracy for the model trained on the unaugmented data. This percentage difference is borne out across a number of test runs. The average precision for the neural net trained on the augmented data set is 85\%, versus 82\% for the model trained on the unaugmented data. The model trained on the augmented data performs better in all respects.

%
%

\begin{table}
\centering
\setlength\tabcolsep{4pt}
\begin{tabular}{r|cccc}
& \textbf{precision} & \textbf{recall} & \textbf{f1-score} & \textbf{support}\\
\hline\\
\textbf{adobe} & 0.75 & 0.60 & 0.67 & 5 \\
\textbf{appsanywhere} & 1.00 & 1.00 & 1.00 & 3 \\
\textbf{blackboard} & 0.91 & 0.91 & 0.91 & 22 \\
\textbf{password} & 0.79 & 0.92 & 0.85 & 12\\
\textbf{wifi} & 0.80 & 0.73 & 0.76 & 11\\
\\
\textbf{accuracy} & & & 0.85 & 53\\
\textbf{macro avg} & 0.85 & 0.83 & 0.84 & 53\\
\end{tabular}
\vspace*{12pt}
\caption{Classification report for neural network with data augmentation.}
\label{tab:classification_report_nn_augmented}
\end{table}

\begin{figure}
\centering
\resizebox{12cm}{!}{\includegraphics{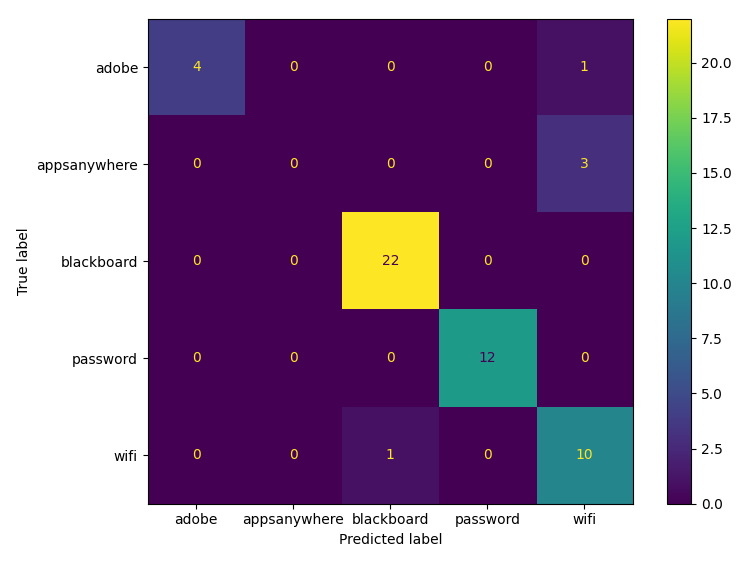}}
\caption{Confusion matrix for decision tree.}
\label{fig:confusion_dt}
\end{figure}

Now we turn our attention to comparing the neural network model with a simpler decision tree model. The confusion matrix for this is shown in figure \ref{fig:confusion_dt}. Looking at the classification report in table \ref{tab:classification_report_dt}, we see that the decision tree is outperforming the neural net in \textit{accuracy}, 91\% compared to 85\%, respectively. However, as we have an imbalanced sample this is a biased metric in favour of the most frequent helpdesk queries about blackboard (a Virtual Learning Environment), as can be seen in the \emph{support} column of the classification reports. We require our email auto-replies to be accurate across the whole range of possible queries so, for this application, a measure that treats each category with equal weight is favoured. We can see by inspection that the decision tree performs particularly badly on the small number of queries about AppsAnywhere (a higher-ed app-store). The precision (the ratio of true positives to the the total number of positive predictions) is zero because there are no true positives at all. Similarly, the recall for this category is also zero, which indicates the complete inability of the classifier to identify the positive samples in this case. This in fact emphasises the well-known high variance (or overfitting) downside to decision tree induction, especially in learning problems with a limited sample size and explicit under-representation of some target categories.   

The (macro) averages of these results are therefore a better indicator of performance in this application, and we can see that these are better for the neural network (with data augmentation) than for the decision tree (a weighted average of scores would produces a similar imbalance as the accuracy, so is not considered here). The results for precision and recall are combined in a \emph{harmonic mean} as the F-measure, or f1 score, which gives us the percentage of positive predictions that were correct. Comparing the (macro) average F-measure for the neural net and decision tree we see that it is greater for the neural net than for the decision tree, 84\% compared to 73\%, despite the fact that the decision tree achieves greater accuracy.

%
%

\begin{table}
\centering
\setlength\tabcolsep{4pt}
\begin{tabular}{r|cccc}
& \textbf{precision} & \textbf{recall} & \textbf{f1-score} & \textbf{support}\\
\hline\\
\textbf{adobe} & 1.00 & 0.80 & 0.89 & 5 \\
\textbf{appsanywhere} & 0.00 & 0.00 & 0.00 & 3 \\
\textbf{blackboard} & 0.96 & 1.00 & 0.98 & 22 \\
\textbf{password} & 1.00 & 1.00 & 1.00 & 12 \\
\textbf{wifi} & 0.71 & 0.91 & 0.80 & 11 \\
\\
\textbf{accuracy} & & & 0.91 & 53 \\
\textbf{macro avg} & 0.73 & 0.74 & 0.73 & 53 \\
\end{tabular}
\vspace*{12pt}
\caption{Classification report for decision tree}
\label{tab:classification_report_dt}
\end{table}

The methods and findings explored here are preliminary, and it is hard to draw general conclusions based on the limited data available. Nonetheless, we see the application of a `light' neural net approach outperforming simpler learning methods on a small data set. Their ability to perform classification tasks in such conditions is further enhanced by data augmentation. In short, the neural net has learned better on all categories whereas the simpler model (the decision tree) focuses primarily on the majority cases.

\section{Conclusion}

The helpdesk system described in this paper produces reliable responses based on classification of the user's initial query. These results provide positive evidence towards the use of data augmentation for small data sets, and demonstrate the improved precision of neural network models, over simpler models such as decision trees. However, they should be treated with caution as the size of dataset is currently too small to draw any strong conclusions, limiting our study in relation to generalisability as well as statistical significance. However, we continue to actively collect data, and are encouraged that tailored auto-replies can provide a useful stop-gap measure until helpdesk agents are able to step in and take over.

\raggedright
\bibliography{autoreply}

\begin{thebibliography}{10}
\providecommand{\url}[1]{\texttt{#1}}
\providecommand{\urlprefix}{URL }
\providecommand{\doi}[1]{https://doi.org/#1}

\bibitem{Alpaydin2014}
Alpaydin, E.: Introduction to machine learning. The MIT Press (2014)

\bibitem{Blei2003}
Blei, D.M., Ng, A.Y., Jordan, M.I.: "latent dirichlet allocation". Journal of
  Machine Learning Research  \textbf{3}((4–5)),  993–1022 (Jan 2003)

\bibitem{Ganegedara2018}
Ganegedara, T.: Natural Language Processing with TensorFlow: Teach language to
  machines using Python's deep learning library. Packt Publishing (2018)

\bibitem{lopez2021}
López, F.: Class imbalance: Random sampling and data augmentation with
  imbalanced-learn.
  \url{https://towardsdatascience.com/class-imbalance-random-sampling-and-data-augmentation-
  with-imbalanced-learn-63f3a92ef04a} (Feb 2021), [Accessed 09 April 2022]

\bibitem{Manning2008}
Manning, C., Raghavan, P., Schütze, H.: Introduction to Information Retrieval.
  Cambridge University Press (2008), available from:
  \url{https://nlp.stanford.edu/IR-book/}

\bibitem{McTear2016}
McTear, M.: The Conversational Interface. Springer International Publishing
  (2016)

\bibitem{Menon2020}
Menon, S.: Stratified sampling in machine learning.
  \url{https://medium.com/analytics-vidhya/stratified-sampling-in-machine-learning-f5112b5b9cfe}
  (Dec 2020), [Accessed 04 April 2022]

\bibitem{Nayak2022}
Nayak, R.: Hands on data augmentation in nlp using nlpaug python library.
  \url{https://medium.com/codex/hands-on-data-augmentation-in-nlp-using-nlpaug-python-library-ad323c22908}
  (Apr 2022), [Accessed 09 April 2022]

\bibitem{Powers2007}
Powers, D.M.W.: Evaluation: From precision, recall and f-factor to roc,
  informedness, markedness \& correlation. Tech. Rep. SIE-07-001, Flinders
  University of South Australia, Adelaide, Australia (2007), available from:
  \url{https://www.researchgate.net/publication/228529307_Evaluation_From_Precision_Recall_and_F-Factor_to_ROC_Informedness_Markedness_Correlation}

\bibitem{Norvig2010}
Russell, S.J., Norvig, P.: Artificial Intelligence: A Modern Approach, Third
  Edition. Prentice Hall (2010)

\bibitem{Shahul2021}
Shahul, E.: Data augmentation in nlp: Best practices from a kaggle master.
  \url{https://neptune.ai/blog/data-augmentation-nlp} (Nov 2021), [Accessed 03
  March 2022]

\bibitem{Wang2017}
Wang, J., Luis, P.: The effectiveness of data augmentation in image
  classification using deep learning, report number: 300. Available from:
  \url{http://cs231n.stanford.edu/reports/2017/pdfs/300.pdf} (2017), [Accessed
  04 April 2022]

\end{thebibliography}
\end{document}